\documentclass[runningheads]{llncs}
\usepackage{graphicx}

\usepackage{subfig}
\usepackage{amsmath,amssymb} % define this before the line numbering.
\usepackage{color}
\usepackage{cite}
\begin{document}
	%===========================================================
	
	\title{Hidden Two-Stream Convolutional Networks for Action Recognition} % Replace your paper's title here
	\titlerunning{Hidden Two-Stream Networks} % Replace an abstracted version of your paper's title here
	
	%===========================================================
	
	\author{Yi Zhu\inst{1} \and
		Zhenzhong Lan \inst{2} \and
		Shawn Newsam\inst{1} \and
		Alexander Hauptmann \inst{2}}
	%
	%Please include author names in full in the paper, 
	%If any authors have names that can be parsed into FirstName LastName in multiple ways, please include the correct parsing, in a comment to the volume editors:
	%\index{Lastnames, Firstnames}
	
	\authorrunning{Y. Zhu et al.} % A shorter version of authors' name
	
	%===========================================================
	
	\institute{University of California at Merced, Merced CA 95343, USA 
		\email{\{yzhu25,snewsam\}@ucmerced.edu}\\
		\and
		Carnegie Mellon University, Pittsburgh PA 15213, USA\\
		\email{\{lanzhzh,alex\}@cs.cmu.edu}}
	
	\maketitle
	
	%===========================================================
	\begin{abstract}
		Analyzing videos of human actions involves understanding the temporal relationships among video frames. State-of-the-art action recognition approaches rely on traditional optical flow estimation methods to pre-compute motion information for CNNs. Such a two-stage approach is computationally expensive, storage demanding, and not end-to-end trainable. In this paper, we present a novel CNN architecture that implicitly captures motion information between adjacent frames. We name our approach hidden two-stream CNNs because it only  takes  raw  video  frames  as input  and  directly  predicts  action classes  without  explicitly computing optical flow. Our end-to-end approach is 10x faster than its two-stage baseline. Experimental results on four challenging action recognition datasets: UCF101, HMDB51, THUMOS14 and ActivityNet v1.2 show that our approach significantly outperforms the previous best real-time approaches. 
		
		\keywords{Action recognition \and Optical flow \and Unsupervised learning.}
	\end{abstract}
	%===========================================================
	\section{Introduction}
	\label{sec:intro}
	
	The field of human action recognition has advanced rapidly over the past few years. We have moved from manually designed features \cite{idtfWang2013,videoDarwin} to learned convolutional neural network (CNN) features \cite{KarpathyCVPR14,c3d2015}; from encoding appearance information to encoding motion information \cite{twostream2014}; and from learning local features to learning global video features \cite{TSN2016,dovf_lan_2017}. The performance has continued to soar higher as we incorporate more of the steps into an end-to-end learning framework. Nevertheless, current state-of-the-art CNN structures still have difficulty in capturing motion information directly from video frames. Instead, traditional local optical flow estimation methods are used to pre-compute motion information for the CNNs \cite{twostream2014}. This two-stage pipeline, first compute optical flow and then learn the mapping from optical flow to action labels, is sub-optimal for the following reasons:
	
	\begin{itemize}
		\item The pre-computation of optical flow is time consuming and storage demanding compared to the CNN step. 
		%	Even when using GPUs, optical flow calculation has been the major computational bottleneck of the current two-stream approaches \cite{twostream2014}, which learn to encode appearance and motion information in two separate CNNs. 
		\item Traditional optical flow estimation is completely independent of the final tasks like action recognition and is therefore potentially sub-optimal. 
		%	Because it is not end-to-end trainable, we cannot extract motion information that is optimal for the desired tasks. 
	\end{itemize}
	
	\begin{figure}[t]
		\centering
		\includegraphics[width=1.0\linewidth]{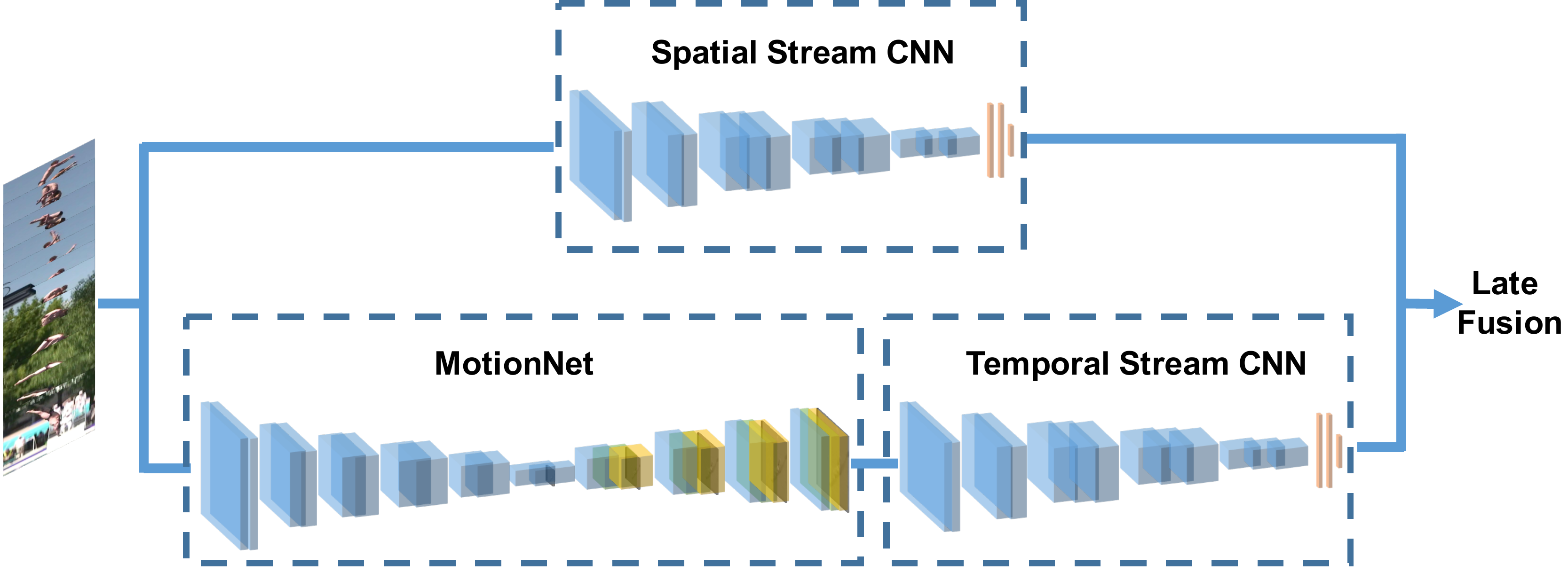}
		\caption{Illustration of proposed hidden two-stream networks. MotionNet takes consecutive video frames as input and estimates motion. Then the temporal stream CNN learns to project the motion information to action labels. Late fusion is performed through the weighted averaging of the prediction scores of the temporal and spatial streams. Both streams are end-to-end trainable.}
		\label{fig:framework}
	\end{figure}
	
	{\color{black} To solve the above problems}, researchers have proposed various methods other than optical flow to capture motion information in videos. For example, new representations like motion vectors \cite{EMV_cvpr16,wu_compressed_action_17} and RGB image difference \cite{TSN2016} or architectures like recurrent neural networks (RNN) \cite{beyondshort2015} and 3D CNNs \cite{c3d2015,qiu_P3D_iccv17,spatiotemporal_du_cvpr2018,spatiotemporal_xie}. However, most of these are not as effective as optical flow for human action recognition\footnote{Detailed comparisons can be found in the supplementary material.}. Therefore, in this paper, we aim to address the above mentioned problems in a more direct way. We adopt the end-to-end CNN approach to learn optical flow so that we can avoid costly computation and storage and obtain task-specific motion representations. 
		% We hope that, by taking consecutive video frames as inputs, our CNNs learn the temporal relationships among pixels and use the relationships to predict action classes.  
		% 	And the task-oriented flow is achieved by end-to-end learning together with the desired task.}
		%Theoretically, given how powerful CNNs are, it would make sense to use them for a low-level task like optical flow estimation. However, in practice, we still face many challenges. Here are the main ones:
		However, we face many challenges to learn such a motion estimation model: 
		
		\begin{itemize}
			\item We need to train the models without supervision. The ground truth flow required for supervised training is usually not available except for limited synthetic data \cite{densenet_flow_icip17,dilate_flow_icip18,guided_flow_17}.  
			\item We need to train our optical flow estimation models from scratch. The models (filters) learned for optical flow estimation tasks are very different from models (filters) learned for other vision tasks \cite{Miao_2018_CVPR,xue2018deep,Miao_2018_AAAI}.
			\item We cannot simply use the traditional optical flow estimation loss functions. We are concerned chiefly with how to learn an optimal motion representation for video action recognition. 
		\end{itemize}
		
		%we choose an unsupervised pre-training approach.
		To address these challenges, we first train a CNN with the goal of generating optical flow from a set of consecutive frames. Through a set of specially designed operators and unsupervised loss functions, our new training step can generate optical flow that is similar to that generated by one of the best traditional methods \cite{TVL1realTime}. As illustrated in {\color{black}the bottom of} Figure \ref{fig:framework}, we call this network MotionNet. Given the MotionNet, we concatenate it with a temporal stream CNN that maps the estimated optical flow to the target action labels. We then fine-tune this stacked temporal stream CNN in an end-to-end manner with the goal of predicting action classes for the input frames. 
		We call our new approach hidden two-stream networks as it implicitly generates motion information for action recognition. Our contributions include: 
		\begin{itemize}
			\item Our method is both computationally and storage efficient. It is around 10x faster than its two-stage baseline, and we do not need to store the pre-computed optical flow.
			\item Our method outperforms previous real-time approaches on four challenging action recognition datasets by a large margin.
			\item The proposed MotionNet is flexible in that it can be directly concatenated with other video action recognition frameworks \cite{beyondshort2015,I3D_Carreira_cvpr17,spatiotemporal_du_cvpr2018,url_cvpr18} to improve their efficiency.
			\item We demonstrate the generalizability of our end-to-end learned optical flow by showing promising results on four optical flow benchmarks without fine-tuning.
		\end{itemize}
		
		\section{Related Work}
		\label{sec:related}
		Significant advances in understanding human activities in video have been made over the past few years. Initially, traditional handcrafted features such as Improved Dense Trajectories (IDT) \cite{idtfWang2013} dominated the field of video analysis for several years. Despite their superior performance, IDT and its improvements are computationally formidable for real applications. CNNs \cite{KarpathyCVPR14,c3d2015}, which are often several orders of magnitude faster than IDTs, performed much worse than IDTs in the beginning. This inferior performance is mostly because CNNs have difficulty in capturing motion information among frames. Later on, two-stream CNNs \cite{twostream2014} addressed this problem by pre-computing the optical flow using traditional optical flow estimation methods \cite{TVL1realTime} and training a separate CNN to encode the pre-computed optical flow. This additional stream (a.k.a., the temporal stream) significantly improved the accuracy of CNNs and finally allowed them to outperform IDTs on several benchmarks. These accuracy improvements indicate the importance of temporal motion information for action recognition as well as the inability of existing CNNs to capture such information. 
		
		However, compared to the CNN, the optical flow calculation is computationally expensive. It is the major speed bottleneck of the current two-stream approaches. As an alternative, Zhang \textit{et al.} \cite{EMV_cvpr16} proposed to use motion vectors to replace the more precise optical flow. This simple improvement brought more than 20x speedup compared to the traditional two-stream approaches. However, this speed improvement came with an equally significant accuracy drop. The encoded motion vectors lack fine structures, and contain noisy and inaccurate motion patterns, leading to much worse accuracy compared to the more precise optical flow \cite{TVL1realTime}. These weaknesses are fundamental and can not be improved. Another more promising approach is to learn to predict optical flow using supervised CNNs, which is closer to our approach. Ng. \textit{et al.} \cite{actionflownet_16} used optical flow calculated by traditional methods as supervision to train a network to predict optical flow. This method avoids the pre-computation of optical flow at inference time and greatly speeds up the process. However, the quality of the optical flow calculated by this approach is limited by the quality of the traditional flow estimation, which again limits its potential on action recognition. Ilg \textit{et al.} \cite{flownet2} use a network trained on synthetic data where ground truth flow exists. The ability of synthetic data to represent the complexity of real data is very limited. Ilg \textit{et al.} \cite{flownet2} actually show that there is a domain gap between real data and synthetic data. To address this gap, they simply grow the synthetic data to narrow the gap. The problem with this solution is that it may not work for other datasets and it is not feasible to do this for all datasets. Our work addresses the optical flow estimation problem in a much more fundamental and promising way. We predict optical flow on-the-fly using CNNs, thus addressing the computation and storage problems. And we perform unsupervised pre-training on real data, thus addressing the domain gap problem. 
		
		Besides the computational problem, traditional optical flow estimation is completely independent of the high-level final tasks like action recognition and is therefore potentially sub-optimal. However, our approach is end-to-end optimized. 
		It is important to distinguish between these two ways of introducing motion information to the encoding CNNs. 
		Although optical flow is currently being used to represent the motion information in the videos, we do not know whether it is an optimal representation. There might be an underlying motion representation that is better than optical flow. In fact, a recent work \cite{xue17toflow} demonstrated that fixed flow estimation is not as good as task-oriented flow for general computer vision tasks. 
		Hence, we believe that our end-to-end learning framework will help us extract better motion representations than traditional optical flow for action recognition. However, for notational convenience, we still refer our learned motion representation as optical flow. 
		
		\section{Hidden Two-Stream Networks}
		\label{sec:method}
		In this section, we describe our proposed hidden two-stream networks in detail. 
		We first introduce our unsupervised network for optical flow estimation along with employed good practices in Section \ref{sec:unsup}. We name it MotionNet. 
		In Section \ref{sec:stacked}, we stack the temporal stream network upon MotionNet to allow end-to-end training. 
		%We call this stacked network the stacked temporal stream CNN. We also experiment with multi-task learning in a branched temporal stream and compare these two design choices. 
		Finally, we introduce the hidden two-stream CNNs in Section \ref{sec:hidden} which combines our stacked temporal stream with a spatial stream.

		\subsection{Unsupervised Optical Flow Learning}
		\label{sec:unsup}
		We treat optical flow estimation as an image reconstruction problem \cite{jasonUnsup2016}. Given a frame pair, we hope to generate the optical flow that allows us to reconstruct one frame from the other. Formally, taking a pair of adjacent frames $I_{1}$ and $I_{2}$ as input, our CNN generates a flow field $V$. Then using the predicted flow field $V$ and $I_{2}$, we get the reconstructed frame $I_{1}^{\prime}$ using backward warping, i.e., $I_{1}^{\prime} = \mathcal{T}[I_{2}, V]$, where $\mathcal{T}$ is the inverse warping function. Our goal is to minimize the photometric error between $I_{1}$ and $I_{1}^{\prime}$. The intuition is that if the estimated flow and the next frame can be used to reconstruct the current frame, then the network should have learned useful representations of the underlying motions.
		
		\noindent \textbf{MotionNet} 
		Our MotionNet is a fully convolutional network, consisting of a contracting part and an expanding part. The contracting part is a stack of convolutional layers and the expanding part is a chain of combined convolutional and deconvolutional layers. 
		The details of our network can be seen in the supplementary material. We describe the challenges and proposed good practices to learn better motion representation for action recognition below. 
		
		First, we design a network that focuses on small displacement motion. For real data such as YouTube videos, we often encounter the problem that foreground motion (human actions of interest) is small, but the background motion (camera motion) is dominant. Thus, we adopt $3\times 3$ kernels throughout the network to detect local, small motions. Besides, we keep the high frequency image details for later stages. Our first two convolutional layers do not use striding. We use strided convolution instead of pooling for image downsampling because pooling is shown to be harmful for dense per-pixel prediction tasks. 
		
		Second, our MotionNet computes multiple losses at multiple scales. Due to the skip connections between the contracting and expanding parts, the intermediate losses can regularize each other and guide earlier layers to converge faster to the final objective. We explore three loss functions that help us to generate better optical flow. These loss functions are as follows. 
		\begin{itemize}
			\item A standard pixelwise reconstruction error function, which is calculated as:
			\begin{equation}
				L_{\text{pixel}} = \frac{1}{ h w} {\sum_{i}^{h} \sum_{j}^{w}} \rho ( I_{1}(i, j) - I_{2}(i+V_{i,j}^{x}, \enskip j+V_{i,j}^{y}) ).
				\label{eq:pixel_loss}
			\end{equation}
			The $V^x$ and $V^y$ are the estimated optical flow in the horizontal and vertical directions. The inverse warping $\mathcal{T}$ is performed using a spatial transformer module \cite{stn_nips15}. Here we use a robust convex error function, the generalized Charbonnier penalty $\rho(x) = (x^{2} + \epsilon^{2})^{\alpha}$,  to reduce the influence of outliers. {\color{black} $h$ and $w$ denote the height and width of images $I_{1}$ and $I_{2}$.}
			
			\item A smoothness loss that addresses the aperture problem that causes ambiguity in estimating motions in non-textured regions. It is calculated as:
			\begin{equation}
				L_{\text{smooth}} = \rho (\nabla V_{x}^{x} ) + \rho ( \nabla V_{y}^{x}) + \rho ( \nabla V_{x}^{y}) + \rho ( \nabla V_{y}^{y}) .
				\label{eq:smoothness_loss}
			\end{equation}
			$\nabla V_{x}^{x}$ and $\nabla V_{y}^{x}$ are the gradients of the estimated flow field $V^{x}$ in each direction. Similarly,  $\nabla V_{x}^{y}$ and $\nabla V_{y}^{y}$ are the gradients of $V^{y}$. The generalized Charbonnier penalty $\rho(x)$ is the same as in the pixelwise loss. 
			
			\item A structural similarity (SSIM) loss function{\color{black}\cite{SSIM_2004}} that helps us to learn the structure of the frames. {\color{black} SSIM is a perceptual quality measure. Given two $K \times K$ image patches $I_{p1}$ and $I_{p2}$, it is calculated as 
				\begin{equation}
					\text{SSIM}(I_{p1}, I_{p2}) = \frac{ (2\mu_{p1}\mu_{p2} + c_{1}) (2\sigma_{p1p2} + c_{2}) }{ (\mu_{p1}^{2} + \mu_{p2}^{2} + c_{1}) (\sigma_{p1}^{2} + \sigma_{p2}^{2} + c_{2}) }. 
					\label{eq:ssim_index}
				\end{equation}
				Here, $\mu_{p1}$ and $\mu_{p2}$ are the mean of image patches $I_{p1}$ and $I_{p2}$, $\sigma_{p1}$ and $\sigma_{p2}$ are the variance of image patches $I_{p1}$ and $I_{p2}$, and $\sigma_{p1p2}$ is the covariance of these two image patches. $c_{1}$ and $c_{2}$ are two constants to stabilize division by a small denominator. In our experiments, $K$ is set to $8$ and $c_{1}$ and $c_{2}$ are 0.0001 and 0.001, respectively. 
				
				In order to compare the similarity between two images $I_{1}$ and $I_{1}^{\prime}$, we adopt a sliding window approach to partition the images into local patches. The stride for the sliding window is set to $8$ in both the horizontal and vertical directions. Hence, our SSIM loss function is defined as: 
				\begin{equation}
					L_{\text{ssim}} = \frac{1}{N} \sum_{n}^{N} ( 1 - \text{SSIM} (I_{1n}, I_{1n}^{\prime}) ) .
					\label{eq:ssim_loss}
				\end{equation}
				
				where \textit{N} is the number of patches we can extract from an image given the sliding stride of $8$, \textit{n} is the patch index. $I_{1n}$ and $I_{1n}^{\prime}$ are two corresponding patches from original image $I_{1}$ and the reconstructed image $I_{1}^{\prime}$.}  
			Our experiments show that this simple strategy significantly improves the quality of our estimated flows. It forces our MotionNet to produce flow fields with clear motion boundaries.
		\end{itemize}
		
		{\color{black} Hence, the loss at each scale $s$} is a weighted sum of the pixelwise reconstruction loss, the piecewise smoothness loss, and the region-based SSIM loss, 
		\begin{equation}
			L_{s} = \lambda_{1} \cdot L_{\text{pixel}} + \lambda_{2} \cdot L_{\text{smooth}}  + \lambda_{3} \cdot L_{\text{ssim}}
			\label{eq:total_unsup_loss}
		\end{equation}
		where $\lambda_{1}$, $\lambda_{2}$, and $\lambda_{3}$ weight the relative importance of the different metrics during training. 
		{\color{black}Since we have predictions at five scales (flow2 to flow6) due to five expansions in the decoder, the overall loss of MotionNet is a weighted sum of loss $L_{s}$: 
			\begin{equation}
				L_{\text{all}} = \sum_{s=1}^{5} \delta_{s} L_{s}
				\label{eq:scale_unsup_loss}
			\end{equation}
			where the $\delta_{s}$ are set to balance the losses at each scale and are numerically of the same order.} We describe how we determine the values of these weights in the supplementary materials.
		
		Third, unsupervised learning of optical flow introduces artifacts in homogeneous regions because the brightness assumption is violated. We insert additional convolutional layers between deconvolutional layers in the expanding part to yield smoother motion estimation.  We also explored other techniques in the literature, like adding flow confidence and multiplying by the original color images \cite{flownet2} during expanding. However, we did not observe any improvements. 
		
		In Section \ref{sec:ablation}, we conduct an ablation study to demonstrate the contributions of each of these strategies. Though our network structure is similar to a concurrent work \cite{flownet2}, MotionNet is fundamentally different from FlowNet2. First, we perform unsupervised learning while \cite{flownet2} performs supervised learning for optical flow prediction. Unsupervised learning allows us to avoid the domain gap between synthetic data and real data. Unsupervised learning also allows us to train the model for target tasks like action recognition in an end-to-end fashion even if the datasets of target applications do not have ground truth optical flow. 
		Second, our network architecture is carefully designed to balance efficiency and accuracy. For example, MotionNet only has one network, while FlowNet2 has 5 similar sub-networks. The model footprints of MotionNet and FlowNet2 \cite{flownet2} are $170$M and $654$M, and the prediction speeds are $370$fps and $25$fps, respectively. We also present an architecture search in the supplementary materials to obtain deep insights in terms of the model trade-off between accuracy and efficiency. 
		%We also propose several effective practices and modifications. These design differences lead to large speed and accuracy differences as will be shown.
		
		\subsection{Projecting Motion Features to Actions}
		\label{sec:stacked}
		Given that MotionNet and the temporal stream are both CNNs, we would like to combine these two modules into one stage and perform end-to-end training. There are multiple ways to design such a combination to project motion features to action labels. Here, we explore two ways, stacking and branching. 
		
		Stacking is the most straightforward approach and just places MotionNet in front of the temporal stream, treating MotionNet as an off-the-shelf flow estimator. Branching is more elegant in terms of architecture design. It uses a single network for both motion feature extraction and action classification. The convolutional features are shared between the two tasks. Due to space limitations, we show in the supplementary materials that stacking is more effective than branching. It achieves better action recognition performance while remaining complementary to the spatial stream. From now on, we choose stacking to project the motion features to action labels.
		
		For stacking, we first need to normalize the estimated flows before feeding them to the encoding CNN. More specifically, as suggested in \cite{twostream2014}, we first clip the motions that are larger than $20$ pixels to $20$ pixels. Then we normalize and quantize the clipped flows to have a range between $0 \sim 255$. We find such a normalization is important for good temporal stream performance and design a new normalization layer for it. 
		
		Second, we need to determine how to fine tune the network, including which loss to use during the fine tuning. We explored different settings. (a) Fixing MotionNet, which means that we do not use the action loss to fine-tune the optical flow estimator. (b) Both MotionNet and the temporal stream CNN are fine-tuned, but only the action categorical loss function is computed. No unsupervised objective (\ref{eq:total_unsup_loss}) is involved. (c) Both MotionNet and the temporal stream CNN are fine-tuned, and all the loss functions are computed. Since motion is largely related to action, we hope to learn better motion estimators by this multi-task way of learning. As will be demonstrated later in Section \ref{sec:results}, model (c) achieves the best action recognition performance. 
		
		Third, we need to capture relatively long-term motion dependencies. We accomplish this by inputting a stack of multiple consecutive flow fields. Simonyan and Zisserman \cite{twostream2014} found that a stack of $10$ flow fields achieves a much higher accuracy than only using a single flow field. To make fair comparison, we also fix the length of our input to be $11$ frames to allow us to generate $10$ optical flows.
		
		\subsection{Hidden Two-Stream Networks}
		\label{sec:hidden}
		
		We also show the results of combining our stacked temporal stream with a spatial stream. These results are important as they are strong indicators of whether our stacked temporal stream indeed learns complementary motion information or just appearance information. 
		
		Following the testing scheme of \cite{twostream2014,wanggoodpractice2015}, we evenly sample $25$ frames/clips for each video. For each frame/clip, we perform $10$x data augmentation by cropping the $4$ corners and $1$ center, flipping them horizontally and averaging the prediction scores (before softmax operation) over all crops of the samples. In the end, we fuse the two streams' scores with a spatial to temporal stream ratio of 1:1.5.

		\section{Experiments}
		\label{sec:experiments}
		
		\subsection{Evaluation Datasets}
		\label{sec:dataset}
		We perform experiments on {\color{black} four} widely used action recognition benchmarks, UCF101 \cite{ucf101}, HMDB51 \cite{hmdb51}, {\color{black} THUMOS14 \cite{THUMOS15} and ActivityNet \cite{activityNet}}.
		UCF101 is composed of realistic action videos from YouTube. It contains $13,320$ video clips distributed among $101$ action classes. HMDB51 includes $6,766$ video clips of $51$ actions extracted from a wide range of sources, such as online videos and movies.
		Both UCF101 and HMDB51 have a standard three-split evaluation protocol and we report the average recognition accuracies over the three splits. {\color{black} THUMOS14 and ActivityNet are large-scale video datasets for action recognition and detection, which contain long untrimmed videos. THUMOS14 has 101 action classes. It includes a training set, validation set, test set and background set. We don't use the background set in our experiments. We use 13,320 training and 1,010 validation videos for training and report the performance on 1,574 test videos. For ActivityNet, we use its 1.2 version which has 100 action classes. Following the standard evaluation split, 4,819 training and 2,383 validation videos are used for training and 2,480 videos for testing. }
		
		\begin{table}[t]
			\begin{center}%The speed includes all stages in inference.
				\caption{Comparison of accuracy and efficiency.
					Top section: Two-stage temporal stream approaches.  
					Middle Section: End-to-end temporal stream approaches.  
					Bottom Section: Two-stream approaches. \label{tab:unsup_results}}
				% \resizebox{0.6\columnwidth}{!}{%
					\begin{tabular}{  c | c | c  }
						\hline
						Method										&    Accuracy (\%)  &    fps \\
						\hline		
						\hline
						TV-L1 \cite{TVL1realTime}							&   $85.65$   &    $14.75$ \\		
						FlowNet \cite{flownet}							&   $55.27$   &    $52.08$ \\		
						FlowNet2 \cite{flownet2}					&   $79.64$   &    $8.05$ \\
						NextFlow \cite{nextflow_16}					&   $72.2$   &    $42.02$ \\
						Enhanced Motion Vectors \cite{EMV_cvpr16}			&   $79.3$   &    $390.7$ \\	
						MotionNet (2 frames)				&   $84.09$   &    $48.54$ \\		
						\hline
						\hline
						ActionFlowNet (2 frames)\cite{actionflownet_16}			&   $70.0$ 	&    $200.0$\\	% 0.0178s
						ActionFlowNet (16 frames)\cite{actionflownet_16}		&   $83.9$ 	&    $-$\\	% 0.0178s
						Stacked Temporal Stream CNN (a)	    					&   $83.76$ 	&    $169.49$\\	% 0.0178s
						Stacked Temporal Stream CNN (b)     					&   $84.04$ 	&    $169.49$\\	
						Stacked Temporal Stream CNN (c)    					&   $84.88$ 	&    $169.49$\\	% 0.0178s
						%Ours Unsup Learned Flow     									&   $82.71$	&    $303.03$ \\	% 0.0033s
						\hline
						\hline
						%Efficient C3D \cite{ali_efficient_c3d_2016}    					&   $85.2$ 	&    $210$\\	% 0.0178s
						Two-Stream CNNs \cite{twostream2014}				&   $88.0$ 	&    $14.3$\\	% 0.0178s
						Very Deep Two-Stream CNNs\cite{wanggoodpractice2015}				&   $\mathbf{90.9}$ 	&    $\mathbf{12.8}$\\	% 0.0178s
						Hidden Two-Stream CNNs (a)    					&   $87.50$ 	&    $120.48$\\	% 0.0178s
						Hidden Two-Stream CNNs (b)    					&   $87.99$ 	&    $120.48$\\	% 0.0178s
						Hidden Two-Stream CNNs (c)    					&   $\mathbf{89.82}$ 	&    $\mathbf{120.48}$\\	% 0.0178s
						\hline
					\end{tabular}
				% } 
			\end{center}
		\end{table} 
		
		\subsection{Results}
		\label{sec:results}
		In this section, we evaluate our proposed framework on the first split of UCF101. We report the accuracy as well as the processing speed of the inference step in frames per second. The results are shown in Table \ref{tab:unsup_results}. The implementation details are in the supplementary materials.
		
		\noindent \textbf{Top section of Table \ref{tab:unsup_results}:} Here we compare the performance of two-stage approaches. {\color{black}By two-stage, we mean optical flow is pre-computed, cached, and then fed to a CNN classifier to project flow to action labels. For fair comparison, our MotionNet here is pre-trained on UCF101, but not fine-tuned using the action classification loss. It only takes frame pairs as input and outputs one flow estimate.}
		The results show that our MotionNet achieves a good balance between accuracy and speed in this setting. 
		
		In terms of accuracy, our unsupervised MotionNet is competitive to TV-L1 while performing much better ($4\% \sim 12\%$ absolute improvement) than other methods of generating flows, including supervised training using synthetic data (FlowNet \cite{flownet} and FlowNet2 \cite{flownet2}), and directly getting flows from compressed videos (Enhanced Motion Vectors \cite{EMV_cvpr16}). These improvements are very significant in datasets like UCF101. 
		In terms of speed, we are also among the best of the CNN based methods and much faster than TV-L1, which is one of the fastest traditional methods.   
		
		%It is worth noting that FlowNet2 is the state-of-the-art CNN flow estimator. It proposes several strategies like stacking multiple networks, using small displacement networks, fusing large and small motion networks etc., to produce accurate and sharp optical flow estimations for different scenarios. However, our MotionNet significantly surpasses the performance of FlowNet2, which indicates the effectiveness of MotionNet for action recognition. 
		
		\noindent  \textbf{Middle section of Table \ref{tab:unsup_results}:} Here we examine the performance of end-to-end CNN based approaches. None of these approaches store intermediate flow information and thus run much faster than the two-stage approaches. If we compare the average running time of these approaches to the two-stage ones, we can see that the time spent on writing and reading intermediate results is almost 3x as much as the time spent on all other steps. Therefore, from an efficiency perspective, it is important to do end-to-end training and predict optical flow on-the-fly. 
		
		ActionFlowNet \cite{actionflownet_16} is what we denote as a branched temporal stream. It is a multi-task learning model to jointly estimate optical flow and recognize actions. The convolutional features are shared which leads to faster speeds. However, even the 16 frames ActionFlowNet performs $1\%$ worse than our stacked temporal stream. Besides, ActionFlowNet uses optical flow from traditional methods as labels to perform supervised training. This indicates that during the training phase, it still needs to cache flow estimates which is computation and storage demanding for large-scale video datasets. Also the algorithm will mimic the failure cases of the classical approaches. 
		
		If we compare the way we fine-tune our stacked temporal stream CNNs, we can see that model (c) where we include all the loss functions to do end-to-end training, is better than the other models including fixing MotionNet weights (model (a)) and only using the action classification loss function (model (b)). These results show that both end-to-end fine-tuning and fine-tuning with unsupervised loss functions are important for stacked temporal stream CNN training.

		\noindent  \textbf{Bottom section of Table \ref{tab:unsup_results}:} Here we compare the performance of two-stream networks by fusing the prediction scores from the temporal stream CNN with the prediction scores from the spatial stream CNN. These comparisons are mainly used to show that stacked temporal stream CNNs indeed learn motion information that is complementary to what is learned in appearance streams. 
		
		The accuracy of the single stream spatial CNN is $80.97\%$. We observe from Table \ref{tab:unsup_results} that significant improvements are achieved by fusing a stacked temporal stream CNN with a spatial stream CNN to create a hidden two-stream CNN. These results show that our stacked temporal stream CNN is able to learn motion information directly from the frames and achieves much better accuracy than spatial stream CNN alone. This observation is true even in the case where we only use the action loss for fine-tuning the whole network (model (b)). This result is significant because it indicates that our unsupervised pre-training indeed finds a better path for CNNs to learn to recognize actions and this path will not be forgotten in the fine-tuning process. If we compare the hidden two-stream CNNs to the stacked temporal stream CNNs, we can see that the gap between model (c) and model (a)/(b) widens.  The reason may be because, without the regularization of the unsupervised loss, the networks start to learn appearance information. Hence they become less complementary to the spatial CNNs. 
		
		Finally, we can see that our models achieve very similar accuracy to the original two-stream CNNs. Among the two representative works we show, Two-Stream CNNs \cite{twostream2014} is the earliest two-stream work and Very Deep Two-Stream CNNs \cite{wanggoodpractice2015} is the one we improve upon. Therefore, Very Deep Two-Stream CNNs \cite{wanggoodpractice2015} is the most comparable work. We can see that our approach is about $1\%$ worse than Very Deep Two-Stream CNNs \cite{wanggoodpractice2015} in terms of accuracy but about 10x faster in terms of speed. 
	
		\begin{table}[t]
			\begin{center}
				\caption{Ablation study of good practices employed in MotionNet. \label{tab:ablation}}
				% \resizebox{0.8\columnwidth}{!}{%
					\begin{tabular}{  c | c | c| c | c | c | c }
						\hline
						Method								  &    Small Disp    &    SSIM    & CDC    &    Smoothness    &    {\color{black}MultiScale} &    Accuracy (\%) \\
						\hline		
						MotionNet							&   $\times$ &    $\times$ &    $\times$ &   $\times$ &   $\times$ & $77.79$       \\		
						MotionNet							&   $\checkmark$ &    $\checkmark$ &    $\checkmark$ &   $\checkmark$ &   $\times$ & {\color{black}$80.63$ }      \\	
						MotionNet							&   $\checkmark$ &    $\checkmark$ &    $\checkmark$ &   $\times$ &   $\checkmark$ & $80.14$       \\		
						MotionNet							&   $\checkmark$ &    $\checkmark$ &    $\times$ &   $\checkmark$ &   $\checkmark$ & $81.25$       \\		
						MotionNet							&   $\checkmark$ &    $\times$ &    $\checkmark$ &   $\checkmark$ &  $\checkmark$ &  $81.58$       \\		
						MotionNet							&   $\times$ &    $\checkmark$ &    $\checkmark$ &   $\checkmark$ &   $\checkmark$ & $82.22$       \\		
						MotionNet							&   $\checkmark$ &    $\checkmark$ &    $\checkmark$ &   $\checkmark$ &   $\checkmark$ & $\mathbf{82.71}$       \\		
						\hline
					\end{tabular}
				% }
			\end{center}
		\end{table} 
		
		\section{Discussion}
		\label{sec:discussion}
		
		\subsection{Ablation Studies for MotionNet} 
		\label{sec:ablation}
		Because of our specially designed loss functions and operators, our proposed MotionNet can produce high quality motion estimates, which allows us to achieve promising action recognition accuracy. Here, we run an ablation study to understand the contributions of these components. The results are shown in Table \ref{tab:ablation}. \textit{Small Disp} indicates using a network that focuses on small displacements. \textit{CDC} means adding an extra convolution between deconvolutions in the expanding part of MotionNet. {\color{black} \textit{MultiScale} indicates computing losses at multiple scales. }
		
		First, we examine the importance of using a network structure that focuses on small displacement motions. We keep the other aspects of the implementation the same, but use a larger kernel size and stride in the beginning of the network. The accuracy drops from $82.71\%$ to $82.22\%$. This drop shows that using smaller kernels with a deeper network indeed helps to detect small motions. 
		
		Second, we examine the importance of adding the SSIM loss. Without SSIM, the action recognition accuracy drops to $81.58\%$. This more than $1\%$ performance drop shows that it is important to focus on discovering the structure of frame pairs. 
% 		Similar observations can be found in \cite{mono_depth_Godard17} for depth estimation.
		
		Third, we examine the effect of removing convolutions between the deconvolutions in the expanding part of MotionNet. This strategy is designed to smooth the motion estimation. As can be seen in Table \ref{tab:ablation}, removing extra convolutions brings a significant performance drop from $82.71\%$ to $81.25\%$.
		
		Fourth, we examine the advantage of incorporating the smoothness objective. Without the smoothness loss, we obtain a much worse result of $80.14\%$. This result shows that our real-world data is very noisy.  Adding smoothness regularization helps to generate smoother flow fields by suppressing noise. This suppression is important for the following temporal stream CNNs to learn better motion representations for action recognition. 
		
		{ \color{black} Fifth, we examine the necessity of computing losses at multiple scales during deconvolution. Without the multi-scale scheme, the action recognition accuracy drops to $80.63\%$. The performance drop shows that it is important to regularize the output at each scale in order to produce the best flow estimation in the end. Otherwise, we found that the intermediate representations during deconvolution may drift to fit the action recognition task, and not predict optical flow. }
		
		Finally, we explore a model that does not employ any of these practices. As expected, the performance is the worst, which is $4.94\%$ lower than our full MotionNet. 
		
		\begin{table}[t]
			\begin{center}%The speed includes all stages in inference.
				\caption{Evaluation of optical flow and action classification. \color{black}{For flow evaluation, lower error is better. For action recognition, higher accuracy is better.} \label{tab:learned_flow_quality}}
				% \resizebox{0.7\columnwidth}{!}{%
					\begin{tabular}{  c | c | c | c | c || c }
						\hline
						Method	    	&   Sintel  &  \color{black}{KITTI2012}  &  \color{black}{KITTI2015}  &  \color{black}{Middlebury}  & UCF101 \\
						\hline
						\hline
						FlowNet2			&   $\mathbf{6.02}$  	&   \color{black}{$\mathbf{1.8}$}  	&   \color{black}{$\mathbf{11.48}$}  	&   \color{black}{$0.52$}  	 									&    $81.97$ \\
						TV-L1 					&   $10.46$   					&   \color{black}{$14.6$} 					&   \color{black}{$47.64$}   							&   \color{black}{$\mathbf{0.45}$}  					&    $\mathbf{85.65}$ \\
						MotionNet			&   $11.93$  					&  \color{black} {$7.5$}  					&   \color{black}{$30.65$}  							&   \color{black}{$0.91$ } 								&    $84.88$ \\
						\hline
					\end{tabular}
				% } 
			\end{center}
		\end{table}

		\subsection{Learned Optical Flow} 
		\label{sec:learned_optical_flow}
		In this section, we systematically investigate the effects of different motion estimation models for action recognition, {\color{black}as well as their flow estimation quality}. We also show some visual examples to discover possible directions for future improvement. 
		Here, we compare three optical flow models: TV-L1, MotionNet and FlowNet2. To quantitatively evaluate the quality of learned flow, we test the three models on {\color{black}four} well received benchmarks, MPI-Sintel, {\color{black}KITTI 2012,  KITTI 2015 and Middlebury}. For action recognition accuracy, we report their performance on UCF101 split1. The results can be seen in Table \ref{tab:learned_flow_quality}. {\color{black}We use EPE (endpoint error) to evaluate MPI-Sintel, KITTI 2012 and Middlebury with lower being better. We use Fl (percentage of optical flow outliers) to evaulate KITTI 2015 with lower being better.  We use classification accuracy to evaluate UCF101 with higher being better.}
		
		{ \color{black} For flow quality, FlowNet2 generally performs better, except on Middlebury because it mostly contains small displacements. Our MotionNet has similar performance to TV-L1 on Sintel and Middlebury, and outperforms TV-L1 on KITTI 2012 and KITTI 2015. The result is encouraging because the KITTI benchmark contains real data (not synthetic), which indicates that the flow estimation from our MotionNet is robust and generalizable. In addition, although FlowNet2 ranks higher on optical flow benchmarks, it performs the worst on action recognition tasks.} This interesting observation means that lower EPE does not always lead to higher action recognition accuracy. This is because EPE is a very simple metric based on L2 distance, which does not consider motion boundary preservation or background motion removal. This is crucial, however, for recognizing complex human actions. 
		
		\begin{figure*}[t]
			\centering
			\includegraphics[width=1.0\linewidth]{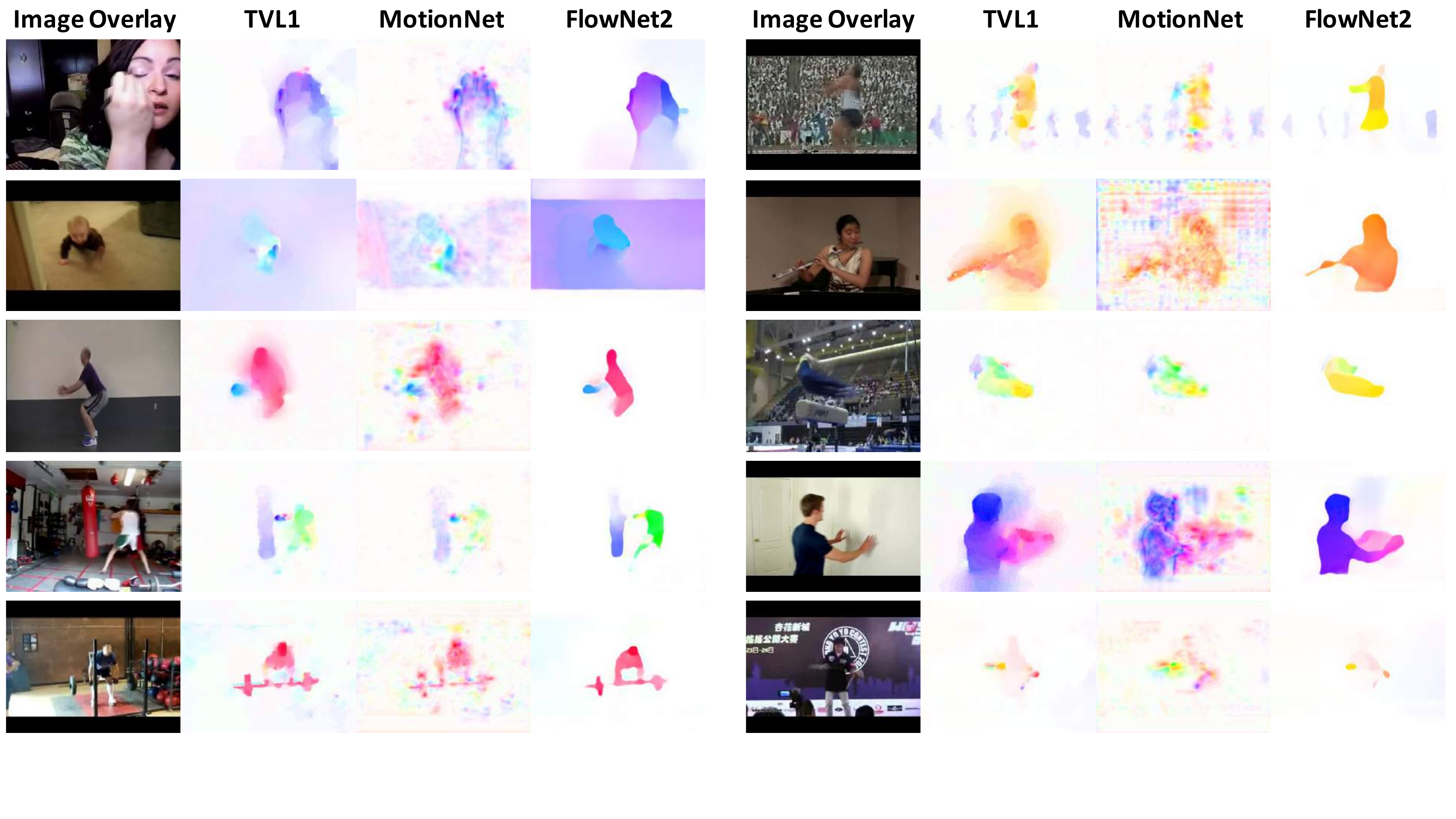}
			\caption{Visual comparisons of estimated flow field from TV-L1, MotionNet and FlowNet2. Left: ApplyEyeMakeup, BabyCrawling, BodyWeightSquats, BoxingPunchingBag and CleanAndJerk. Right: Hammering, PlayingFlute, PommelHorse, WallPushups and YoYo. This figure is best viewed in color.}
			\label{fig:learned_flow}
		\end{figure*}
		
		We also show some visual samples in Figure \ref{fig:learned_flow} to help understand the effect of the quality of estimated flow fields for action recognition. The color scheme follows the standard flow field color coding in \cite{flownet2}. In general, the estimated flow fields from all three models look reasonable. 
		MotionNet has lots of background noise compared to TV-L1 due to its global learning. This maybe the reason why it performs worse than TV-L1 for action recognition. FlowNet2 has very crisp motion boundaries, fine structures and smoothness in homogeneous regions. It is indeed a good flow estimator in terms of both EPE and visual inspection. However, it achieves much worse results for action recognition, $3.5\%$ lower than TV-L1 and $2.9\%$ lower than our MotionNet. Thus, which motion representation is best for action recognition remains an open question. 
		
		% Given these observations, it is very important to introduce an appropriate metric to evaluate the quality of motion estimates for video understanding tasks.

		\section{Comparison to State-of-the-Art Real-Time Approaches}
		\label{sec:sota}
		In this section, we compare our proposed method to recent real-time state-of-the-art approaches as shown in Table \ref{tab:sota}\footnote{In general, the requirement for real-time processing is $25$ fps. We also compare to other non real-time approaches in the supplementary materials.}. Among all real-time methods, our hidden two-stream networks achieves the highest accuracy on the {\color{black}four} benchmarks. We also show the flexibility of our MotionNet by concatenating it to temporal streams with different backbone CNN architectures, e.g., VGG16 \cite{wanggoodpractice2015}, TSN \cite{TSN2016} and I3D \cite{I3D_Carreira_cvpr17}. With deeper networks, we can achieve higher recognition accuracy and still be real-time. We are $6.1\%$ better on UCF101, $14.2\%$ better on HMDB51, {\color{black} $8.5\%$ better on THUMOS14 and $7.8\%$ better on ActivityNet} than the previous state-of-the-art. This indicates that {\color{black} our stacked end-to-end learning framework can implicitly learn better motion representations than motion vectors \cite{kantorov2014,EMV_cvpr16} and RGB differences \cite{TSN2016} with respect to the task of action recognition.}

		\begin{table}[t]
			\begin{center}
				\caption{Comparison to state-of-the-art real-time approaches {\color{black} on four benchmarks with respect to mean classification accuracy.} $^{*}$ indicates results from our implementation.  \label{tab:sota}}
				% \resizebox{1.0\columnwidth}{!}{%
					\begin{tabular}{  c | c | c | c | c }
						\hline
						Method																	&    UCF101(\%)  &    HMDB51(\%) &    {\color{black} THUMOS14(\%)}  &   {\color{black}  ActivityNet(\%)} \\
						\hline		
						\hline
						MV + FV \cite{kantorov2014}	    &   $78.5$   &    $46.7$ &   $-$   &    $-$ \\
						EMV  \cite{EMV_cvpr16}					&   $80.2$   &    $-$ 		&  {\color{black} $41.6$}  &    $-$ \\
						C3D (1 Net)  \cite{c3d2015}											&   $82.3$   &    $49.7^{*}$ 	&   {\color{black}$54.6$}   &   {\color{black}  $74.1$} \\
						%			ActionFlowNet (2 frames)  \cite{actionflownet_16}		    &   $70.0$ 	&    	$42.6$		&   $-$   &    $-$ \\	
						ActionFlowNet \cite{actionflownet_16}		    &   $83.9$ 	&    	$56.4$		&   $51.3^{*}$   &    $68.8^{*}$ \\
						
						%			C3D (3 Net)  \cite{c3d2015}											&   $85.2$   &    $-$ 		&   $-$   &    $-$ \\
						RGB + EMV  \cite{EMV_cvpr16}					&   $86.4$   &    $-$ 		&  {\color{black}$61.5$}  &    $-$ \\
						% 	Two-Stream 3DNet Initial  \cite{ali_efficient_c3d_2016}    					&   $85.2$ 	&    	$-$		&   $78.5$   &    $46.7$ \\
						% 	Two-Stream 3DNet Mid  \cite{ali_efficient_c3d_2016}    	&   $87.0$ 	&    	$-$\\
						3DNet  \cite{ali_efficient_c3d_2016}    &   $90.2$ 	&    	$-$		&   $-$   &    $-$ \\
						%			RGB Diff    \cite{TSN2016}                       &   $83.0$ 	&    	$-$		&   $-$   &    $-$ \\
						%			RGB + RGB Diff     \cite{TSN2016}                       &   $86.8$ 	&    	$-$		&   $-$   &    $-$ \\	
						RGB Diff (TSN)    \cite{TSN2016}             &   $91.0$ 	&    	$64.5^{*}$		&   $71.9^{*}$  &   $83.0^{*}$ \\
						\hline
						\hline
						%			Hidden two-stream CNNs (Tiny-MotionNet )    	&   $88.7$ 					&    $58.9$							&  {\color{black}$63.2$}							&    {\color{black}$71.3$} \\
						Ours (VGG16)    				&   $90.3$ 					&    $60.5$							&   {\color{black}$66.7$}				&   {\color{black}$77.8$}\\
						Ours (TSN) 						&   $93.2$ 		&    $66.8$		& $74.5$   			&   $87.9$ \\	
						Ours (I3D) 						&   $\mathbf{97.1}$ 		&    $\mathbf{78.7}$		&  {\color{black}$\mathbf{80.6}$}   			&   {\color{black}$\mathbf{91.2}$} \\
						% \hline
						% \hline
						% Two-Stream \cite{twostream2014}				&   $88.0$ 	&    $58.0$\\	% 0.0178s
						% Bilen \textit{et al.} \cite{dynamicImage}						&   $89.1$ 	&    $65.2$\\	% 0.0178s
						% Very Deep Two-Stream \cite{wanggoodpractice2015}				&   $91.4$ & $62.9$ \\
						% Wang \textit{et al.} \cite{wang2016actions}				&   $92.4$ 	&    $62.0$\\	% 0.0178s
						% fusion \textit{et al.} \cite{wang2016actions}				&   $92.4$ 	&    $62.0$\\	% 0.0178s
						% keyvolume \textit{et al.} \cite{wang2016actions}				&   $92.4$ 	&    $62.0$\\	% 0.0178s
						% Zhu \textit{et al.} \cite{depth2action}					&   $93.0$ 	&    $68.2$\\	% 0.0178s
						% TSN \cite{TSN2016}									&   $94.2$ 	&    $69.4$\\	% 0.0178s
						% Qiu \textit{et al.} \cite{deep_quantization_16}                       & $95.2$ & $-$  \\
						% Lan \textit{et al.} \cite{dovf_lan_2017}                              & $95.3$ & $75.0$  \\
						% Diba \textit{et al.} \cite{diba_tle_2016}							&   $95.6$ 	&    $71.1$\\	% 0.0178s
						\hline
					\end{tabular}
				% } 
			\end{center}
		\end{table} 
		
		\section{Conclusion}
		\label{sec:conclusion}
		
		We have proposed a new framework called hidden two-stream networks to recognize human actions in video. It addresses the problem of capturing the temporal relationships among video frames which the current CNN architectures have difficulty with. Different from the current common practice of using traditional local optical flow estimation methods to pre-compute the motion information for CNNs, we use an unsupervised pre-training approach. Our MotionNet is computationally efficient and end-to-end trainable. It is flexible and can be directly applied in other frameworks for various video understanding applications. Experimental results on four challenging benchmarks demonstrate the effectiveness of our approach.
		
		In the future, we would like to improve our hidden two-stream networks in the following directions. First, we would like to improve our optical flow prediction based on the observation that the smoothness loss has significant impact on the quality of the motion estimations for action recognition. Second, we would like to incorporate other best practices that improve the overall performance of the networks. For example, joint training of the two streams instead of a simple late fusion. Third, it would be interesting to see how addressing the false label assignment problem can help improve our overall performance. Finally, removing global camera motion and partial occlusion within the CNN framework would be helpful for both optical flow estimation and action recognition. 
	
	\noindent \textbf{Acknowledgement} We gratefully acknowledge the support of NVIDIA Corporation through the donation of the Titan Xp GPUs used in this work.
	
		\bibliographystyle{splncs04}
	\bibliography{journalBIB}

\end{document}